\documentclass[10pt,twocolumn,letterpaper]{article}

\usepackage[pagenumbers]{cvpr} 

\definecolor{cvprblue}{rgb}{0.21,0.49,0.74}
\usepackage[pagebackref,breaklinks,colorlinks,allcolors=cvprblue]{hyperref}

\def\paperID{7812} 
\def\confName{CVPR}
\def\confYear{2025}

\title{InteractionMap: Improving Online Vectorized HDMap Construction with Interaction}

\author{Kuang Wu \qquad
Chuan Yang \thanks{Corresponding at: ycscience86@gmail.com} \qquad
Zhanbin Li \qquad \\
Langge Technology
}

\begin{document}
\maketitle
\begin{abstract}
    Vectorized high-definition (HD) maps are essential for an autonomous driving system. Recently, state-of-the-art map vectorization methods are mainly based on DETR-like framework to generate HD maps in an end-to-end manner. In this paper, we propose InteractionMap, which improves previous map vectorization methods by fully leveraging local-to-global information interaction in both time and space. Firstly, we explore enhancing DETR-like detectors by explicit position relation prior from point-level to instance-level, since map elements contain strong shape priors. Secondly, we propose a key-frame-based hierarchical temporal fusion module, which interacts temporal information from local to global. Lastly, the separate classification branch and regression branch lead to the problem of misalignment in the output distribution. We interact semantic information with geometric information by introducing a novel geometric-aware classification loss in optimization and a geometric-aware matching cost in label assignment. InteractionMap achieves state-of-the-art performance on both nuScenes and Argoverse2 benchmarks.
\end{abstract}
\section{Introduction}\label{sec:intro}

High-Definition (HD) maps are designed for high-precision autonomous driving, which contain instance-level vectorized representation such as lane divider, road boundaries, pedestrian crossing, \etc. The rich semantic information of road topology and traffic rules is important for the navigation of autonomous driving (AD). HD maps are traditionally constructed offline using LiDAR SLAM-based methods~\cite{zhang2014loam, shan2018lego} with high maintenance costs, complex pipelines, and notable localization errors. In addition, manual annotation and map updating rely heavily on human labor and time demands.

In recent years, more research endeavors have shifted towards deep-learning-based methods that construct vectorized HD maps around the ego-vehicle at runtime with onboard sensors~\cite{li2022hdmapnet, liao2022maptr, li2023lanesegnet}. With the development of bird's eye view (BEV) representation, recent approaches~\cite{chen2022efficient, zhou2022cross, hu2021fiery,li2022bevformer, philion2020lift, li2022hdmapnet, roddick2020predicting} treat the semantic map learning task as a semantic segmentation problem. However, rasterized maps generated by these methods lack vectorized instance-level information, which is essential to downstream tasks, such as motion forecasting and motion planning~\cite{gao2020vectornet, liang2020learning}. To overcome the limitations of segmentation-based methods, approaches VectorMapNet~\cite{liu2023vectormapnet},  MapTR~\cite{liao2022maptr}, MapTRv2~\cite{liao2023maptrv2} predict point sets to construct end-to-end vectorized local HD maps using transformer~\cite{vaswani2017attention} decoders inspired by DETR~\cite{carion2020end}. 

\begin{figure}[t]
    \centering
    \includegraphics[width=0.95\linewidth]{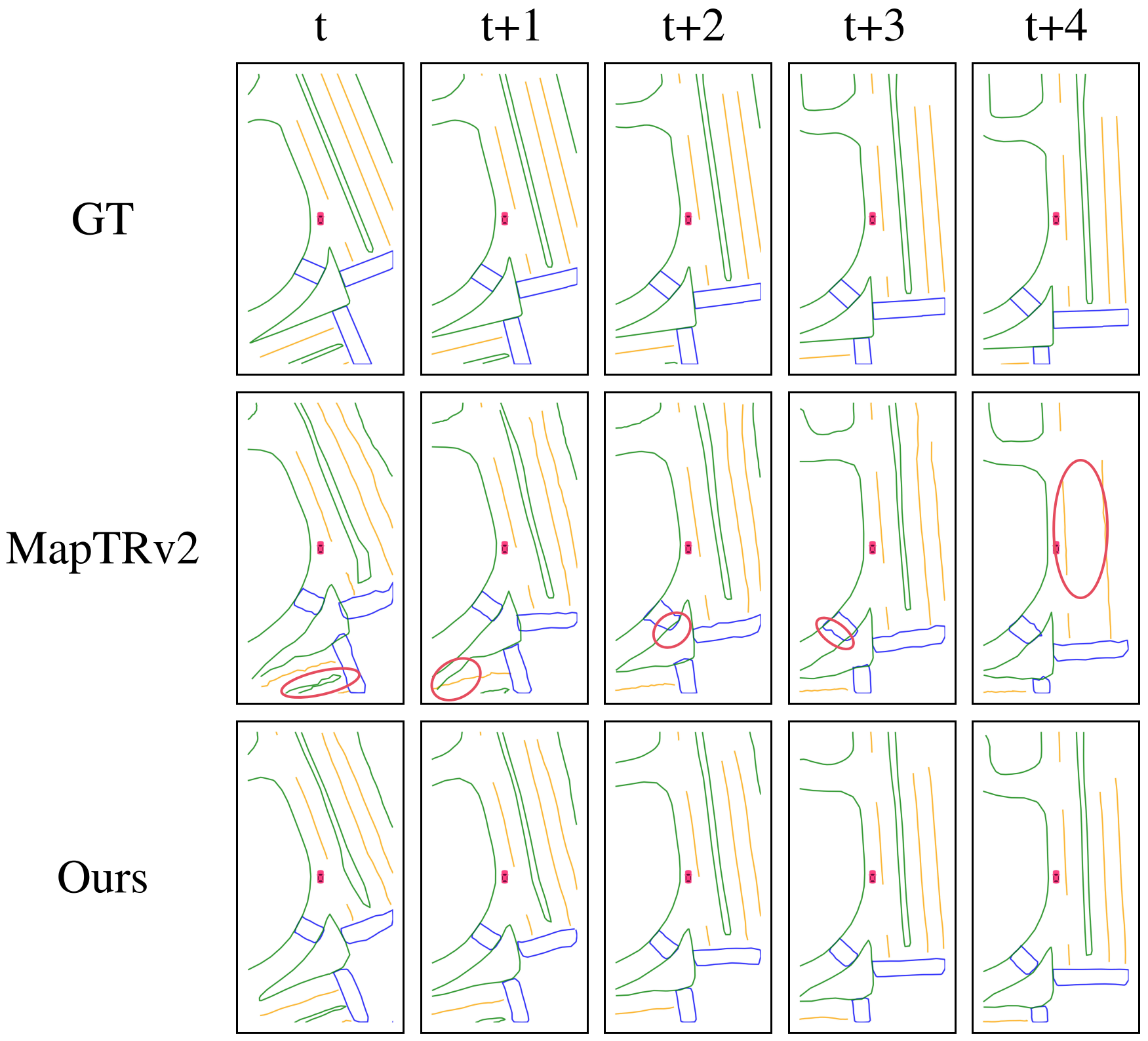}
    \caption{Visual comparison between MapTRv2~\cite{liao2023maptrv2} and our improved results. Our method effectively eliminates error map elements, leading to better precision and stability.}
    \label{fig:compare}
  \end{figure}

As shown in Figure~\ref{fig:compare}, the representation of point sets has a limited model capability of instance-level information. Moreover, predicting directly from a single frame input is challenging in complex scenarios, due to temporal inconsistency between frames. To address these issues, we propose InteractionMap, which consists of Relation Embedding Module (REM), Temporal Fusion Module (TFM), and Geometry-aware Alignment Module (GAM).
REM establishes position relation representation among feature embeddings at point level and instance level, improving the precision of map construction by explicit geometric priors. TFM facilitates long-range temporal associations by key-frame-based streaming strategy, a hierarchical temporal fusion from local to global. It enhances the stability of map elements among different frames in some complex scenarios, such as map element occlusion caused by moving vehicles. GAM solves the misalignment problem caused by inconsistency of the predictions between the classification score and the instance points precision, \eg.\ a prediction with a high classification score and relatively low localization quality (large Chamfer distance).

Our main contributions can be summarized as follows:
\begin{itemize}
    \item We introduce an explicit position relation embedding method from point to instance, efficiently employing progressive interaction of point-wise information and instance-wise information.
    \item We propose a novel key-frame-based hierarchical streaming strategy, which fully leverages long-range temporal information.
    \item We present a geometry-aware classification loss and a geometry-aware matching cost to overcome the misalignment problem of classification and localization output. 
    \end{itemize}

\section{Related Work}\label{sec:relatedwork}

\begin{figure*}[t]
    \begin{center}
    \centering
    \includegraphics[width=1.0\textwidth]{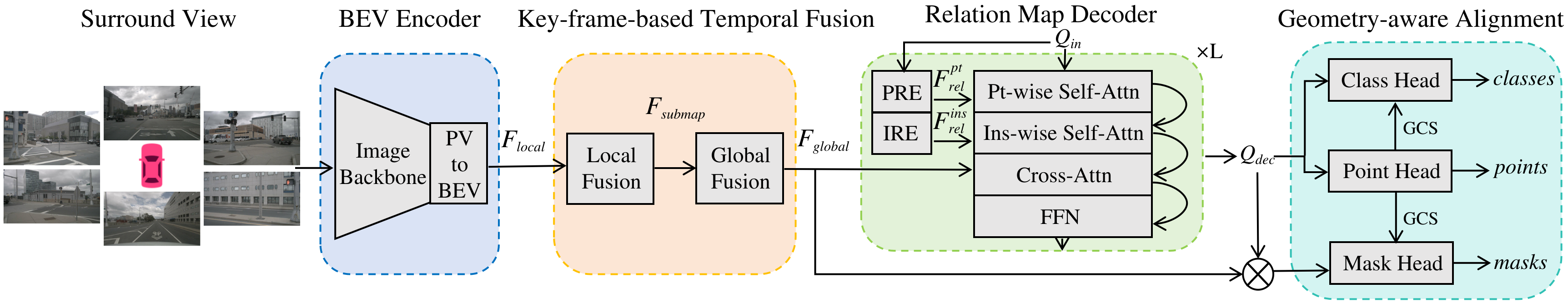}
    \vspace{-6mm}
    \caption{\textbf{Overview of InteractionMap framework.} InteractionMap mainly consists of four components: (1) BEV encoder transforms sensor input to a unified BEV representation; (2) Key-frame-based temporal fusion module leverage temporal information from local to global; (3) Relation map decoder utilizes relation embedding in point-level and instance-level; (4) Geometry-aware alignment module is designed to solve the misalignment problem of classification and position output.}
    \label{fig:ppl}
    \vspace{-6mm}
    \end{center}
    \end{figure*} 

\subsection{HD Map Construction}
With the development of view transformation from perspective-view (PV) to bird-eye-view (BEV) methods~\cite{ma2024vision, li2023delving}, HD map construction is formulated as a segmentation task~\cite{chen2022efficient, zhou2022cross, hu2021fiery,li2022bevformer, philion2020lift, liu2023vision, liu2023bevfusion, pan2023baeformer, zhang2022beverse} based on sensor observations on board.
HDMapNet~\cite{li2022hdmapnet} builds a vectorized HD map using semantic segmentation, clustering, and post-processing. VectorMapNet~\cite{gao2020vectornet} is the first end-to-end framework that utilizes transformers~\cite{vaswani2017attention}, in a two-stage coarse-to-fine manner. However, the auto-regressive model of VectorMapNet leads to a long training schedule. MapTR~\cite{liao2022maptr} adopts a one-stage transformer approach based on~\cite{carion2020end, zhu2020deformable, li2022bevformer} with a permutation-equivalent point set modeling. The evolved version MapTRv2~\cite{liao2023maptrv2} adds auxiliary headers, decoupled self-attention in the decoder and one-to-many matching strategy, leading to a large improvement. Instead of a point set representation, BeMapNet~\cite{qiao2023end} utilizes an instance-level representation with a piecewise Bezier head. PivotNet~\cite{ding2023pivotnet} converts point-level representation to instance-level representation using the point-to-line mask module. MapQR~\cite{liu2025leveraging} implicitly encodes point-level queries within instance-level queries and embeds query positions like Conditional DETR~\cite{meng2021conditional} and DAB-DETR~\cite{liu2022dab}.
MapVR~\cite{zhang2024online} generates a vectorized map with differentiable rasterization that provides instance-level segmentation supervision. MGMap~\cite{liu2024mgmap} employs mask-guided features to refine an instance representation with enhanced feature detail.
In contrast to the above methods, we introduce more information interaction to improve the reliability and stability of map construction in complex scenarios.

\subsection{Online Temporal Fusion}
The temporal fusion strategy is effective for the 3D object detection task, such as BEVFormer~\cite{li2022bevformer}, SOLOFusion~\cite{park2022time}, StreamPETR~\cite{wang2023exploring}, VideoBEV~\cite{han2024exploring} and Sparse4D v2~\cite{lin2023sparse4d}. HDMapNet~\cite{li2022hdmapnet} employs max pooling to fuse temporal information directly into the BEV feature map. StreamMapNet~\cite{yuan2024streammapnet} proposes query propagation and BEV feature map fusion. SQD-MapNet~\cite{wang2024stream} designs a stream query denoising approach. MapTracker~\cite{chen2025maptracker} enhances query propagation and BEV feature map fusion using a tracking strategy. The streaming strategy facilitates longer temporal association as the propagated hidden states encode all historical information. However, a temporal encoder such as the gated recurrent unit (GRU)~\cite{chung2014empirical} may still face the problem of forgetting due to limited capacity in complex outdoor environments. For example, occlusion of moving vehicles may lead to temporal fusion of polluted BEV feature map. The stacking strategy may integrate features from specific previous frames, offering flexibility in fusion of long-range information. In this paper, we propose a key-frame-based streaming strategy, leveraging local and global fusion capabilities.

\subsection{Map Element Interaction}
ADMap~\cite{hu2024admap} explores point-order relationships between and within instances through a cascading approach. InsightMapper~\cite{xu2023insightmapper} performs inner-instance feature aggregation by an additional masked inner-instance self-attention module. GeMap~\cite{zhang2023online} designs a masked decoupled self-attention to handle the shapes and relations of the instance independently. HoMap~\cite{cai2024homap} introduces high-order modeling to capture the correlations between instances using high-order statistics. HIMap~\cite{zhou2024himap} proposed a hybrid framework to learn and interact with information at the point-level and the element-level.
As proved in~\cite{hu2018relation}, modeling relations between objects is beneficial to object recognition. In state-of-the-art methods~\cite{hou2024relation, gao2024ease}, relationship of queries improves DETR-like models. In this paper, we explore the interaction of map elements between their geometry and appearance feature by our explicit position relation embedding.

\subsection{Classification-Localization Alignment}
The classification head and the localization head of object detection are implemented separately by two branches in parallel, causing inconsistency in the output distribution. This misalignment problem has been well researched in the 2D object detection field~\cite{li2020generalized, zhang2021varifocalnet}. However, it is overlooked by the map element detectors~\cite{liao2022maptr, liao2023maptrv2, ding2023pivotnet, qiao2023end, zhang2024online, liu2025leveraging}. 
In this paper, we create a geometry-aware classification loss, making use of focal loss~\cite{ross2017focal} with a geometry-aware target in foreground candidates. Moreover, we present a geometry-aware matching cost to suppress the matching of background candidates.

\section{Method}\label{sec:method}
Figure~\ref{fig:ppl} illustrates the framework of our proposed InteractionMap. Firstly, BEV features are extracted from multi-view images by the BEV encoder, which consists of a backbone~\cite{he2016deep} to extract multi-scale image features, and a view transformation module~\cite{li2022bevformer, liu2023bevfusion} to encode image features into a BEV representation $\mathcal{F}_{local}$. Secondly, key-frame-based temporal fusion leverages long-range temporal information by local fusion and global fusion (Section~\ref{ssec:tempfusion}). Subsequently, the relation map decoder utilizes point-wise relation embedding (PRE) and instance-wise relation embedding (IRE) for better interaction between queries (Section~\ref{ssec:relation}). Finally, by introducing a geometric-aware classification loss and a geometric-aware matching cost, the geometry-aware alignment leverages interaction between semantic information and geometry information (Section~\ref{ssec:geoalign}).

\subsection{Key-frame-based Temporal Fusion}\label{ssec:tempfusion}
Compared to the single-frame map element prediction method, there are mainly two types of temporal fusion methods, namely stacking-based strategy and streaming-based strategy. The streaming strategy leverages all historical information by hidden state encoding. However, due to limited memory capacity, the streaming strategy has limited performance in long-range perception. The stacking strategy fuse features from specific previous frames, offering flexibility in integration of long-range information. The computational cost is linearly related to the number of fused frames.

\begin{figure}[t]
    \centering
    \includegraphics[width=0.95\linewidth]{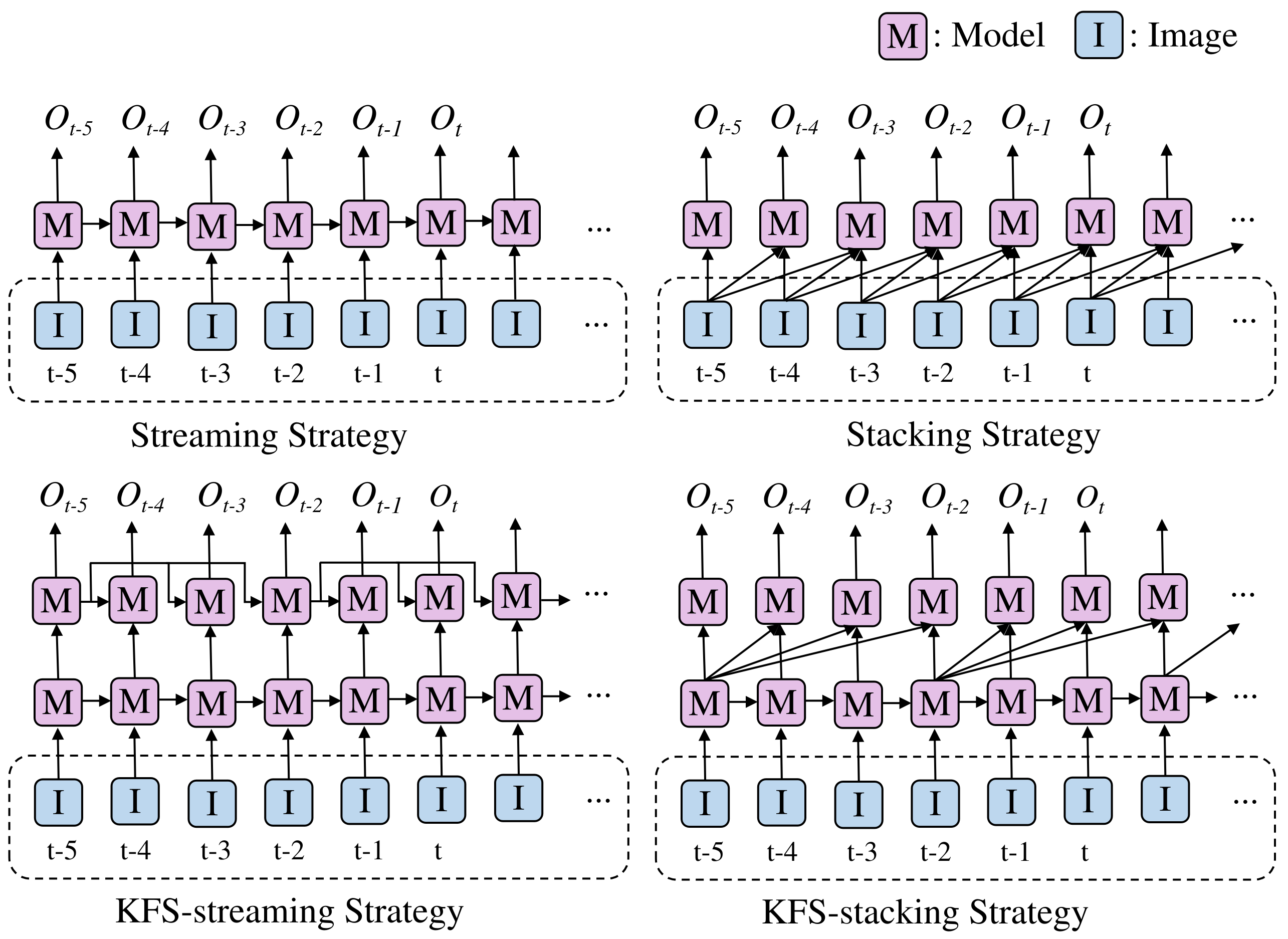}
    \caption{Temporal fusion strategy.}
    \label{fig:streamstack}
    \end{figure}

We propose a key-frame-based streaming (KFS) strategy to address these problems. The well-known key-frame-based strategy is widely used in robot navigation and mapping, such as sparse mapping~\cite{klein2007parallel, mur2015orb} and dense mapping~\cite{zhou2013elastic, dai2017bundlefusion}. Online reconstruction methods are unstable in long range, but they are accurate in the local regime. The input frames are divided into local frame segments. The submaps are reconstructed by integration of local frames, and then the global map is reconstructed by integration of submaps. We introduce two types of KFS strategy, namely KFS-streaming and KFS-stacking, as shown in Figure~\ref{fig:streamstack}. Firstly, the BEV feature embedding submaps are generated by a streaming-base temporal encoder at the bottom level, which keeps the stability and temporal consistency of predictions cross latest frames. Secondly, the long-range BEV feature embedding is generated by a streaming-based or stacking-based temporal encoder at the top level. Figure~\ref{fig:streamstack} illustrates the KFS strategy that merges one previous submap with an interval of two.

\subsubsection{Local BEV Fusion}
We warp the memorized BEV feature map from the previous frame to the current frame recurrently based on the ego vehicle's relative pose. Then we employ a GRU~\cite{chung2014empirical} to fuse current BEV feature map $\mathcal{F}^{t}_{local}$ and the warped memorized BEV feature map $\Tilde{\mathcal{F}}^{t-1}_{submap}$ into a single BEV feature embedding submap candidate $\mathcal{F}^{t}_{submap}$.

\begin{equation}
    \Tilde{\mathcal{F}}^{t-1}_{submap} = {Warp}(\mathcal{F}^{t-1}_{submap}, \boldsymbol{T})
    \end{equation}
\begin{equation}
    \mathcal{F}^{t}_{submap} = ResBlock({LN}(GRU(\Tilde{\mathcal{F}}^{t-1}_{submap}, \mathcal{F}^{t}_{local})))
    \end{equation}
Where $ResBlock$ is a residual-based convolution block with the shortcut connection~\cite{he2016deep}. $LN$ is a layer normalization operation. $\boldsymbol{T}$ denotes a standard $4 \times 4$ transformation matrix between the coordinate systems of two frames.

\subsubsection{Global BEV Fusion}
Our KFS strategy selects a BEV feature embedding as a submap using a specific distance stride $d_{stride}$, instead of a specific temporal interval customly.
\\
{\bf KFS-streaming strategy}
The previous selected global BEV feature is warped to the current frame. The warped global BEV feature and the current BEV feature map are then fused to create a global BEV feature map.
\begin{equation}
    \Tilde{\mathcal{F}}^{pre}_{global} = {Warp}(\mathcal{F}^{pre}_{global}, \boldsymbol{T})
    \end{equation}
\begin{equation}
    {\mathcal{F}}^{t}_{global} = {LN}(GRU(\Tilde{\mathcal{F}}^{pre}_{global}, \mathcal{F}^{t}_{submap}))
    \end{equation}
{\bf KFS-stacking strategy}
We warp $N_{pre}$ selected submaps to the current frame. Following concatenation and residual-based convolution~\cite{he2016deep}, we obtain the global BEV feature.
\begin{equation}
    \Tilde{\mathcal{F}}^{t_k}_{submap} = {Warp}(\mathcal{F}^{t_k}_{submap}, \boldsymbol{T})
    \end{equation}
\begin{equation}
    {\mathcal{F}}^{t}_{global} = ResBlock(Concat(\Tilde{F}_{submap}, \mathcal{F}^{t}_{submap}))
    \end{equation}
Here, $\Tilde{\mathcal{F}}^{t_k}_{submap}$ and $t_k$ represent the selected BEV feature submap and its corresponding index. The collection of selected BEV feature submaps is denoted as $\Tilde{F}_{submap} = {\{\Tilde{\mathcal{F}}^{t_k}_{submap}\}}^{N_{pre}}_{k=1}$.
During the training period, the previous $N_{pre}$ submaps are randomly selected within a range of $N_{pre} \times d_{stride}$.

\subsection{Relation map decoder}\label{ssec:relation}
Previous work has shown the effectiveness of interactions for map element detectors~\cite{hu2024admap,zhou2024himap}. In contrast to these approaches, we directly construct explicit relation priors by incorporating relation embedding into the self-attention module of the decoder.
\subsubsection{Relation Embedding}
Here is the self-attention in the decoder of a DETR-like framework.
\begin{equation}
    Q_{self\_attn} = Softmax(\frac{Q K^T}{\sqrt{d_k}})V
    \end{equation}
Here $d_k$ is the dimension of the keys. $Q, K, V$ are feature embeddings generated by linear projections. $Q = Linear_q(Q_{in})$. $K = Linear_k(Q_{in})$. $V = Linear_v(Q_{in})$

We incorporate relation embeddings into the self-attention model, inspired by~\cite{hu2018relation}.
\vspace{-1mm}
\begin{equation}
    Q_{self\_attn} = Softmax(\mathcal{F}_{rel} + \frac{Q K^T}{\sqrt{d_k}})V
    \end{equation}
Here $\mathcal{F}_{rel}$ represents the relation embedding. In this paper, point-wise relation embedding and instance-wise relation embedding are added to the decoupled self-attention model at the point level and the instance level, respectively, employing attention refinement of information interaction.
\subsubsection{Point-wise Relation Embedding}
The point-wise relation encoder represents the high-dimensional point relation embedding as an explicit geometry prior. The point-wise relation embedding is generated by normalized point coordinates and edge directions.
\begin{equation}
    Rel_{pt}(i,j) = [\log(x_i - x_j + 1), \log(y_i - y_j + 1)]
    \end{equation}
\begin{equation}
    Rel_{dir}(i,j) = cosine\_sim(e_i, e_{i+1}) - cosine\_sim(e_j, e_{j+1})
    \end{equation}
Here $x_i$, $x_j$ are normalized point coordinates. $e_i$ and $e_{i+1}$, $e_j$ and $e_{j+1}$ are adjacent edges. The cosine similarity is denoted by $cosin\_sim$.
The point-wise relation embedding is unbiased, since both $Rel_{pt}(i,j)$ and $Rel_{dir}(i,j)$ are equal to 0 when $i=j$.

The point-wise relation embedding $\mathcal{F}^{pt}_{rel}$ is illustrated by the following equation:
\begin{equation}
    \mathcal{F}^{pt}_{rel} = ReLU(Linear(SPE(Concat(Rel_{pt}, Rel_{dir}))))
    \end{equation}
Firstly, the point-wise relation embedding is calculated by concatenation of $Rel_{pt}$ and $Rel_{dir}$. Then the embedding is transformed to high dimension by sine positional encoding (SPE). Lastly, the embedding is transformed by a linear projection and a ReLU function.

\subsubsection{Instance-wise Relation Embedding}
The instance-wise relation embedding is generated by classification scores and chamfer distance of points.
\begin{equation}
    Rel_s(i,j) = sgn(s_i - s_j)
    \end{equation}
\begin{equation}
    Rel_{sd}(i,j) = sgn(s_i - s_j) \cdot Chamfer(I_i, I_j)
    \end{equation}
Here, $sgn$ is the sign function. $s_i$, $s_j$ are classification scores of the $i^{th}$ and $j^{th}$ candidates. $Chamfer(I_i, I_j)$ is the chamfer distance between points of instance $i$ and instance $j$. The equation of $Rel_s(i,j)$ establishes a simple classification ranking relation among pairs of instances. Then we add the position relation of the map elements to the instance-wise relation embedding $\mathcal{F}^{ins}_{rel}$ by the chamfer distance.
\begin{equation}
    \mathcal{F}^{ins}_{rel} = ReLU(Linear(SPE(Rel_{sd})))
    \end{equation}
The instance-wise relation embedding is unbiased, since $Rel_{sd}(i,j)$ is equal to 0 when $i=j$.

\subsection{Geometry-aware Alignment}\label{ssec:geoalign}
Current work, such as MapTR~\cite{liao2022maptr} uses the classification branch and the regression branch to predict semantic labels and geometric localization separately. MapTR uses the focal loss~\cite{ross2017focal} as the classification loss and the point-to-point L1 loss as the regression loss.
Since classification and regression losses are independent, the quality of these two branches is inconsistent, leading to the well-known misalignment problem.
In this paper, we introduce the geometry-aware focal loss and the geometry-aware focal cost to overcome this problem.
\subsubsection{Geometry-aware Focal Loss}
Inspired by IoU-aware Focal Loss~\cite{li2020generalized, zhang2021varifocalnet} in traditional object detection, we design the novel Geometry-aware Focal Loss (GFL). Unlike focal loss, we treat positives and negatives asymmetrically.
\begin{equation}\label{eq:geofocalloss}
    \begin{split}
    \mathcal{L}_{GFL} = \sum\limits_{i=1}^{N_{pos}} s_{geo_i} BCE(s_{geo_i}, p_i) + \sum\limits_{j=1}^{N_{neg}} \alpha p_j^\gamma BCE(p_j, 0)
    \end{split}
    \end{equation}
where $p_i$ and $p_j$ are the predicted probability of the $i^{th}$ and $j^{th}$ candidates, $s_{geo} \in [0,1]$ is an instance Geometry-aware Classification Score (GCS). Here we define three types of GCS, including $s_{p2p}$, $s_{dir}$ and $s_{giou}$.
\begin{equation}\label{eq:p2pscore}
    s_{p2p} = 1 - \dfrac{1}{2 N_p} \sum\limits_{i=1}^{N_p} D_{Manhattan}(p_{src_{i}}, p_{tgt_{i}})
    \end{equation}
\begin{equation}\label{eq:dirscore}
    s_{dir} = 0.5 + \dfrac{1}{2 N_e} \sum\limits_{i=1}^{N_e} cosine\_sim(e_{src_{i}}, e_{tgt_{i}})
    \end{equation}
\begin{equation}\label{eq:giouscore}
    s_{giou} = 0.5 + 0.5 \cdot GIoU(B_{src}, B_{tgt})
    \end{equation}
where $s_{p2p} \in [0,1]$ is a normalized point-to-point L1 score, $s_{dir} \in [0,1]$ is an edge direction score calculated by cosine similarity and $s_{giou}$
$\in [0,1]$ is a normalized GIoU~\cite{rezatofighi2019generalized} score of candidate bounding box $B_{src}$ and assigned GT $B_{tgt}$. The bounding box is computed by minimum enclosing box of instance points. $N_p$ and $N_e$ are the numbers of points and edges. $p_{src}$ and $p_{tgt}$ are normalized points of candidates and assigned GT.\@ $e_{src}$ and $e_{tgt}$ are edges of candidates and assigned GT.\@
\subsubsection{Geometry-aware Focal Cost}
The loss function $\mathcal{L}_{GFL}$ supervises the classification scores with geometric metrics GCS $s_{geo_i}$.
And $s_{geo_i}$ accurately selects these high-quality candidates, which leads to better map vectorization performance.
Following the spirit of GFL, we make similar modifications to the focal cost, which is a key component of label assignment.
The novel matching cost, geometry-aware focal cost (GFC) is as follows.
\begin{equation}\label{eq:geofocalcost}
    \begin{split}
    \mathcal{C}_{GFC}(i,j) = s_{geo_i} BCE(s_{geo_i}, p_i) - \alpha p_i^\gamma BCE(1-p_i, 1)
    \end{split}
    \end{equation}
where $\mathcal{C}_{GFC}(i,j)$ is the geometry-aware focal cost for the $i^{th}$ prediction and the $j^{th}$ ground truth.
The GFC is used as a modulated function to suppress candidates with inaccurate prediction localization and promote candidates with an accurate position.

\subsection{Training Loss}
{\bf Auxiliary BEV Supervision.}
We introduce a query-based instance segmentation (QIS) module using an instance mask prediction branch. Inspired by~\cite{li2023mask}, we make use of the query embeddings $Q_{dec}$ from Transformer to dot-product the BEV feature embedding map $\mathcal{F}_{bev}$ to obtain instance-wise binary masks $m$ with the sigmoid function $\sigma()$.
\begin{equation}\label{eq:maskdino}
    \hat{M}_{ins} = \sigma(MLP(Q_{dec}) \cdot \mathcal{F}_{bev})
    \end{equation}
For the instance segmentation branch, we extend the mask focal loss and the mask focal cost to mask geometry-aware focal loss (MGFL) and mask geometry-aware focal cost (MGFC) as well, with the same ideal of GFL.\@
\begin{equation}
    \mathcal{L}_{seg} = \lambda_{mgf} \mathcal{L}_{mgf}(\hat{M}_{ins}, M_{ins}) + \lambda_{dice} \mathcal{L}_{dice}(\hat{M}_{ins}, M_{ins})
    \end{equation}
Instance masks $\hat{M}_{ins}$ are supervised by instance mask annotations $M_{ins}$, using the mask geometry-aware focal loss $\mathcal{L}_{mgf}$ and the dice loss $\mathcal{L}_{dice}$.
\\
{\bf Overall Loss.}
The total loss $L$ used by InteractionMap is as follow:
\begin{equation}
    \mathcal{L} = \mathcal{L}_{det} + \mathcal{L}_{seg} + \mathcal{L}_{aux}
    \end{equation}
where $\mathcal{L}_{det}$ is the loss term for the detection task. $\mathcal{L}_{seg}$ is the loss term for the segmentation task. $\mathcal{L}_{aux}$ is the auxiliary loss term.

The detection loss consists of classification loss $\mathcal{L}_{cls}$, point-to-point distance loss $\mathcal{L}_{p2p}$, edge direction loss $\mathcal{L}_{dir}$.
\begin{equation}
    \mathcal{L}_{det} = \lambda_{cls} \mathcal{L}_{cls} + \lambda_{p2p} \mathcal{L}_{p2p} + \lambda_{dir} \mathcal{L}_{dir}
    \end{equation}
The loss term $\mathcal{L}_{cls}$ is the geometry-aware focal loss for the classification of instances. The loss term $\mathcal{L}_{p2p}$ is the smooth L1 loss for the regression of vectorized map elements. And the loss term $\mathcal{L}_{dir}$ is the loss of cosine similarity between adjacent edges.

\begin{equation}
    \mathcal{L}_{aux} = \lambda_{depth} \mathcal{L}_{depth} + \lambda_{PVSeg} \mathcal{L}_{PVSeg}
    \end{equation}
The auxiliary loss term $\mathcal{L}_{aux}$ is composed of the depth-map prediction loss $\mathcal{L}_{depth}$ and image semantic segmentation loss $\mathcal{L}_{PVSeg}$.

\section{Experiment}\label{sec:experiment}

\subsection{Experimental Settings}

\subsubsection{Datasets}
{\bf nuScenes Datasets.} There are 1000 scenes of 4 locations in nuScenes datasets. Each scene contains 20s of RGB images from six cameras, point clouds from LiDAR sweeps, and a 3D vectorized map. The train set contains 700 scenes with 28130 samples and validation set contains 150 scenes with 6019 samples, respectively. Following previous works~\cite{liao2022maptr,liao2023maptrv2}, we mainly focus on three categories of map elements, including lane divider, pedestrian crossing and road boundary.
\\
{\bf Argoverse2 Datasets.} Argoverse2 dataset provide 1000 scenes in 6 cities, each lasting around 15 seconds. Each log includes 7 RGB images from surrounding cameras and point cloud from LiDAR sweeps. Follow previous methods~\cite{liao2022maptr,liao2023maptrv2}, 700 scenes are used for training, and 150 scenes are used for validation. For a fair comparison, we report results on its validation set and focus on the same three map categories as the nuScene dataset.
\subsubsection{Evaluation Metrics} For a fair comparison, we follow the standard metric used in previous works~\cite{li2022hdmapnet, liu2023vectormapnet, liao2022maptr, liao2023maptrv2}. With the ego-coordination as the center, the perception ranges are $[-15m, 15m]$ for the X-axis and $[-30m,30m]$ for the Y-axis. We adopt the mean Average Precision (mAP) to evaluate the map vectorization quality based on the chamfer distance. A candidate is considered as True-Positive (TP) only if the chamfer distance to ground truth is less than certain thresholds ($\tau \in T, T = {0.5, 1.0, 1.5}$). The final AP metric is calculated by average across all thresholds and all classes.
\subsubsection{Implementation Details}
Following previous method~\cite{liao2022maptr,liao2023maptrv2}, we use ResNet50~\cite{he2016deep} as the backbone for RGB images. All models are trained with 8 A800 GPUs. The default batch size is 32. The initial learning rate is set to $6 \times 10^{-4}$ with cosine decay. The nuScenes images resolution is $1600 \times 900$. We resize the images with 0.5 ratio. For the Argoverse2 dataset, the image resolutions are $2048 \times 1550$ and $1550 \times 2048$. We pad the images to $2048 \times 2048$, and then resize the images with 0.3 ratio. We define the size of BEV feature as $H \times W$ of $200 \times 100$, and the size of each BEV grid as 0.3 meters. The default numbers of instance queries, point queries, and decoder layers are 100, 20, 6, respectively. The model is trained on 24, 110 epochs on the nuScenes and Argoverse2 datasets. We randomly divide each training sequence into 3 splits to generate more diverse data sequences. Inspired by SOLOFusion~\cite{park2022time}, we train the initial $N_{init}$ epochs with single-frame input to stabilize multi-frame straining. $N_{init} = 0.5 N_{total}$. For loss weights, $\lambda_{cls}=2$, $\lambda_{p2p}=4$, $\lambda_{dir}=0.005$, $\lambda_{mgf}=30$, $\lambda_{dice}=3$, $\lambda_{depth}=3$, $\lambda_{PVSeg}=2$. For temporal fusion, $N_{pre}=4$, $d_{stride}=5m$.

\subsection{Performance Comparison}

\begin{table*}[t]
    \begin{center}
    \centering
    \resizebox{0.74\textwidth}{!}{
    \begin{tabular}{@{}cccc|cccc@{}}
    \toprule
    Method & Backbone & Epochs & Temporal & AP$_{ped}$ & AP$_{div}$ & AP$_{bou}$ & mAP \\
    \midrule

    MapTR & R50 & 24 && 46.3 & 51.5 & 53.1 & 50.3 \\
    MapVR & R50 & 24 && 47.7 & 54.4 & 51.4 & 51.2 \\
    PivotNet & R50 & 30 && 53.8 & 55.8 & 59.6 & 57.4 \\
    BeMapNet & R50 & 30 && 57.7 & 62.3 & 59.4 & 59.8 \\
    MapTRv2 & R50 & 24 && 59.8 & 62.4 & 62.4 & 61.5 \\

    StreamMapNet & R50 & 30 & \checkmark& 61.7 & 66.3 & 62.1 & 63.4 \\
    MGMap & R50 & 24 && 61.8 & 65.0 & 67.5 & 64.8 \\
    SQD-MapNet & R50 & 24 & \checkmark& 63.6 & 66.6 & 64.8 & 65.0 \\
    MapQR & R50 & 24 && 63.4 & 68.0 & 67.7 & 66.4 \\
    HIMap & R50 & 30 && 62.6 & 68.4 & 69.1 & 66.7 \\
    HRMapNet-MapTRv2 & R50 & 24 & \checkmark& 65.8 & 67.4 & 68.5 & 67.2 \\
    \textbf{InteractionMap-R} & R50 & 24 & \checkmark& 71.3 & 70.8 & 72.8 & 71.6 \\
    \textbf{InteractionMap-C} & R50 & 24 & \checkmark& 69.7 & 72.7 & 73.0 & 71.8 \\

    \midrule

    VectorMapNet & R50 & 110+ft && 42.5 & 51.4 & 44.1 & 46.0 \\
    MapTR & R50 & 110 && 56.2 & 59.8 & 60.1 & 58.7 \\
    MapVR & R50 & 110 && 55.0 & 61.8 & 59.4 & 58.8 \\
    BeMapNet & R50 & 110 && 62.6 & 66.7 & 65.1 & 64.8 \\
    MapTRv2 & R50 & 110 && 68.1 & 68.3 & 69.7 & 68.7 \\
    MapQR & R50 & 110 && 70.1 & 74.4 & 73.2 & 72.6 \\
    HRMapNet-MapTRv2 & R50 & 110 & \checkmark& 72.0 & 72.9 & 75.8 & 73.6 \\
    HIMap & R50 & 110 && 71.3 & 75.0 & 74.7 & 73.7 \\
    \textbf{InteractionMap-R} & R50 & 110 & \checkmark& 75.2 & 75.0 & 76.1 & 75.5 \\
    \textbf{InteractionMap-C} & R50 & 110 & \checkmark& 75.3 & 75.6 & 77.0 & 76.0 \\

    \bottomrule
    \end{tabular}
    }
    \end{center}
    \vspace{-3mm}
    \caption{Comparison with SOTA methods on the nuScenes validation set at $60m\times30m$ perception range. ``EB0'', ``EB4'', ``R50'' correspond to the backbones Efficient-B0, Efficient-B4~\cite{tan2019efficientnet}, ResNet50~\cite{he2016deep}. ``ft'' means the two-stage fine-tune strategy. ``C'' means KFS-stacking strategy and ``R'' means KFS-streaming strategy.}
    \vspace{-4mm}
    \label{tab:nus}
    \end{table*}

\begin{table}[t]
    \begin{center}
    \centering
    \renewcommand\tabcolsep{3pt}
    \resizebox{0.47\textwidth}{!}{
    \begin{tabular}{@{}cc|cccc@{}}
    \toprule
    Method & Dim & AP$_{ped}$ & AP$_{div}$ & AP$_{bou}$ & mAP \\
    \midrule

    StreamMapNet & 2d & 62.0 & 59.5 & 63.0 & 61.5 \\
    SQD-MapNet & 2d & 64.9 & 60.2 & 64.9 & 63.3 \\
    MapTRv2 & 2d & 60.0 & 68.7 & 64.2 & 64.3 \\
    HRMapNet-MapTRv2 & 2d & 65.1 & 71.4 & 68.6 & 68.3 \\
    \textbf{InteractionMap-C} & 2d & 68.5 & 77.9 & 74.6 & 73.7 \\
    \textbf{InteractionMap-R} & 2d & 70.1 & 77.0 & 74.0 & 73.7 \\

    \bottomrule
    \end{tabular}
    }
    \end{center}
    \vspace{-3mm}
    \caption{Comparison with SOTA methods on the Argoverse2 validation set at $60m\times30m$ perception range with a 30-epoch training schedule, at a 2Hz sampling frequency.}
    \vspace{-6mm}
    \label{tab:av2_ep30}
    \end{table}

\begin{table}[t]
    \begin{center}
    \centering
    \renewcommand\tabcolsep{5pt}
    \resizebox{0.47\textwidth}{!}{
    \begin{tabular}{@{}cc|cccc@{}}
    \toprule
    Method & Dim & AP$_{ped}$ & AP$_{div}$ & AP$_{bou}$ & mAP \\
    \midrule

    MapTRv2 & 2d & 63.6 & 71.5 & 67.4 & 67.5 \\
    MapQR & 2d & 64.3 & 72.3 & 68.1 & 68.2 \\
    HIMap & 2d & 69.0 & 69.5 & 70.3 & 69.6 \\
    \textbf{InteractionMap-C} & 2d & 69.8 & 78.6 & 74.8 & 74.4 \\
    \textbf{InteractionMap-R} & 2d & 70.8 & 78.1 & 74.9 & 74.6 \\

    \midrule

    MapTRv2 & 3d & 60.7 & 68.9 & 64.5 & 64.7 \\
    MapQR & 3d & 60.1 & 71.2 & 66.2 & 65.9 \\
    HIMap & 3d & 66.7 & 68.3 & 70.3 & 68.4 \\
    \textbf{InteractionMap-C} & 3d & 66.6 & 75.6 & 72.7 & 71.6 \\
    \textbf{InteractionMap-R} & 3d & 67.7 & 75.5 & 73.1 & 72.1 \\

    \bottomrule
    \end{tabular}
    }
    \end{center}
    \vspace{-3mm}
    \caption{Comparison with SOTA methods on the Argoverse2 validation set at $60m\times30m$ perception range with a 6-epoch training schedule, at a 10Hz sampling frequency.}
    \vspace{-1mm}
    \label{tab:av2_ep6}
    \end{table}

\begin{table}[t]
    \begin{center}
    \centering
    \renewcommand\tabcolsep{12pt}
    \resizebox{0.47\textwidth}{!}{
    \begin{tabular}{@{}c|ccc@{}}
    \toprule
    Method & FPS & Params(MB) & mAP \\
    \midrule

    MapTRv2 & 14.9 & 40.6 & 61.5 \\
    \textbf{InteractionMap-C} & 11.0 & 116.7 & 71.8 \\
    \textbf{InteractionMap-R} & 12.1 & 49.9 & 71.6 \\

    \bottomrule
    \end{tabular}
    }
    \end{center}
    \vspace{-3mm}
    \caption{Comparison with SOTA methods on the nuScenes validation set. FPS is measured on a single A800 GPU.}
    \vspace{-5mm}
    \label{tab:fps_mem}
    \end{table}

{\bf Performance on nuScenes Dataset.}
As shown in Table~\ref{tab:nus}, our methods show consistent improvements over baselines. By integrating our method with MapTRv2~\cite{liao2023maptrv2}, we achieve 71.6(+10.1), 71.8(+10.3) mAP with 24-epoch training schedules and 75.5(+6.8), 76.0(+7.3) mAP with 110-epoch training schedules, respectively. InteractionMap establishes a new state-of-the-art performance, exhibiting significant improvements over existing methods with comparable inference speed, as shown in Table~\ref{tab:fps_mem}.
\\
{\bf Performance on Argoverse2 Dataset.}
Argoverse2 dataset contains 3D map elements, including height information. For fair comparison, the sampling frequency is set to 2Hz and 10Hz, respectively. The amount of training data at a 10Hz sampling frequency is 5 times as many as that at a 2Hz sampling frequency. As demonstrated in Table~\ref{tab:av2_ep30} and Table~\ref{tab:av2_ep6}, InteractionMap surpasses all state-of-the-art methods in both 2D and 3D vectorized map perception on Argoverse2 dataset.

\subsection{Ablation Study}\label{ssec:ablation}
We examine the efficacy of each component of InteractionMap through ablation studies, utilizing the nuScenes dataset. Initially, we build a simple baseline base on MapTR~\cite{liao2022maptr} with the LSS encoder without auxiliary depth-map supervision. The default numbers of instances, points per instance, are 50 and 20, respectively.
\\
{\bf Contributions of Main Components.}
Table~\ref{tab:ablation} demonstrates the impact of each component of InteractionMap. When adding relation embedding into the baseline model, there is a substantial improvement of 4.5 mAP.\@ Furthermore, the addition of geometry alignment results in a significant 7.5 increase in mAP, emphasizing its effectiveness in boosting performance. Lastly, key-frame-based temporal fusion offers 5.0 and 5.1 additional increases in mAP, indicating that the inclusion of long-range hierarchical streaming strategy significantly enhances the performance.
\begin{table}[t]
    \centering
    \renewcommand\tabcolsep{7pt}
    \resizebox{0.47\textwidth}{!}{
    \begin{tabular}{@{}ccc|cccc@{}}
    \toprule
    REM & GAM & TFM & AP$_{ped}$ & AP$_{div}$ & AP$_{bou}$ & mAP \\
    \midrule
    &&& 46.7 & 50.5 & 53.5 & 50.2 \\
    \checkmark&&& 50.1 & 56.6 & 57.3 & 54.7 \\
    &\checkmark&& 57.1 & 63.8 & 63.7 & 61.5 \\
    &&C& 50.9 & 60.0 & 61.9 & 57.6 \\
    &&R& 50.6 & 61.3 & 60.3 & 57.4 \\
    \checkmark&\checkmark&& 58.6 & 65.1 & 62.9 & 62.2 \\
    \checkmark&\checkmark&C& 64.2 & 68.6 & 68.8 & 67.2 \\
    \checkmark&\checkmark&R& 64.1 & 68.1 & 69.8 & 67.3 \\
    \bottomrule
    \end{tabular}
    }
    \caption{Ablation study of each component, including Relation Embedding Module (REM), Temporal Fusion Module (TFM), Geometry-aware Alignment Module (GAM). ``C'' means KFS-stacking strategy and ``R'' means KFS-streaming strategy.}
    \vspace{-1mm}
    \label{tab:ablation}
    \end{table}

\\
{\bf Ablation on Relation Embedding.}
In Table~\ref{tab:relation}, we explore the effect of relation embedding in point-level and instance-level. Adding the point-wise relation embedding of point coordinates and edge direction results in 2.1 and 0.9 improvements in mAP over the baseline, respectively. The instance-wise relation embedding leverages the classification confidence and points coordinates, leading to enhancements of 1.1 and 3.1 mAP, respectively. The combination of relation embedding in point-level and instance-level enhances performance by 2.8 mAP.\@ Finally, we introduce a alternating training strategy to randomly turn off one of these two components, leading the overall design achieves 54.7(+4.5) mAP.\@
\begin{table}[t]
    \centering
    \renewcommand\tabcolsep{4pt}
    \resizebox{0.47\textwidth}{!}{
    \begin{tabular}{@{}ccccc|cccc@{}}
    \toprule
    Pt$_{xy}$ & Pt$_{dir}$ & Ins$_{s}$ & Ins$_{sd}$ & PP & AP$_{ped}$ & AP$_{div}$ & AP$_{bou}$ & mAP \\
    \midrule
    &&&&& 46.7 & 50.5 & 53.5 & 50.2 \\
    \checkmark&&&&& 47.8 & 52.7 & 56.3 & 52.3 \\
    &\checkmark&&&& 47.0 & 52.4 & 53.9 & 51.1 \\
    \checkmark&\checkmark&&&& 47.8 & 55.6 & 56.4 & 53.3 \\
    &&\checkmark&&& 46.9 & 52.1 & 54.8 & 51.3 \\
    &&&\checkmark&& 47.9 & 55.7 & 56.4 & 53.3 \\
    \checkmark&\checkmark&&\checkmark&& 47.6 & 55.9 & 55.5 & 53.0 \\
    \checkmark&\checkmark&&\checkmark&\checkmark& 50.1 & 56.6 & 57.3 & 54.7 \\
    \bottomrule
    \end{tabular}
    }
    \caption{Ablation study of relation embedding by employing point-wise coordinate embedding Pt$_{xy}$, point-wise edge direction embedding Pt$_{dir}$, instance-wise classification score embedding Ins$_{s}$, instance-wise classifier score and chamfer distance embedding Ins$_{sd}$ and alternating training strategy of relation embedding, namely Ping-Pong (PP).}
    \vspace{-1mm}
    \label{tab:relation}
    \end{table}

\\
{\bf Ablation on Key-frame-based Temporal Fusion.}
To investigate the effectiveness of the key-frame-based streaming strategy, we conduct ablation experiments to compare it with the baseline (without temporal fusion) and the streaming strategy, in Table~\ref{tab:stream}. Employing the streaming strategy results in 3.8 mAP over the baseline. Furthermore, when the KFS-stacking strategy is introduced, the mAP increased by 7.4, reaching 57.6. Additionally, the inclusion of KFS-streaming strategy leads to an improvement of 7.2 mAP.\@
\begin{table}[t]
    \begin{center}
    \centering
    \renewcommand\tabcolsep{10pt}
    \resizebox{0.47\textwidth}{!}{
    \begin{tabular}{@{}c|cccc@{}}
    \toprule
    Method & AP$_{ped}$ & AP$_{div}$ & AP$_{bou}$ & mAP \\
    \midrule

    Baseline & 46.7 & 50.5 & 53.5 & 50.2 \\
    Streaming & 48.6 & 54.7 & 58.8 & 54.0 \\
    KFS-stacking & 50.9 & 60.0 & 61.9 & 57.6 \\
    KFS-streaming & 50.6 & 61.3 & 60.3 & 57.4 \\

    \bottomrule
    \end{tabular}
    }
    \end{center}
    \vspace{-3mm}
    \caption{Ablation study of streaming strategy, KFS-streaming strategy and KFS-stacking strategy.}
    \vspace{-3mm}
    \label{tab:stream}
    \end{table}

\\
{\bf Ablation on Geometry-aware Alignment.}
As shown in Table~\ref{tab:align}, geometry-aware alignment comprised five main components: query-based instance segmentation (QIS), geometry-aware focal loss (GFL), geometry-aware focal cost (GFC), mask geometry-aware focal loss (MGFL) and mask geometry-aware focal cost (MGFC). The QIS leverages geometry priors by instance segmentation mask supervision. Compared to the baseline, the QIS leads to an improvement of 6.4 mAP.\@ The GFL encourages predictions with high localization metrics to obtain a better classification score, with an improvement of 3.1 mAP compared to mask focal loss. The GFC suppresses the matching cost of candidates with an inaccurate prediction position, which brings an improvement of 0.9 mAP.\@ For the segmentation branch, the result shows that the position-supervised segmentation loss MGFL and the position-modulated segmentation cost MGFC improve the final results, with 0.4 mAP and 0.5 mAP gains, individually.

\begin{table}[t]
    \centering
    \renewcommand\tabcolsep{2pt}
    \resizebox{0.47\textwidth}{!}{
    \begin{tabular}{@{}ccccc|cccc@{}}
    \toprule
    QIS & GFL &GFC & MGFL & MGFC & AP$_{ped}$ & AP$_{div}$ & AP$_{bou}$ & mAP \\
    \midrule
    &&&&& 46.7 & 50.5 & 53.5 & 50.2 \\
    \checkmark&&&&& 51.7 & 58.7 & 59.4 & 56.6 \\
    \checkmark&\checkmark&&&& 55.1 & 62.0 & 62.0 & 59.7 \\
    \checkmark&\checkmark&\checkmark&&& 55.8 & 63.8 & 62.2 & 60.6 \\
    \checkmark&\checkmark&\checkmark&\checkmark&& 57.0 & 63.6 & 62.3 & 61.0 \\
    \checkmark&\checkmark&\checkmark&\checkmark&\checkmark& 57.1 & 63.8 & 63.7 & 61.5 \\
    \bottomrule
    \end{tabular}
    }
    \caption{Ablation study of geometry-aware alignment by investigating the performance of query-based instance segmentation (QIS), geometry-aware focal loss (GFL), geometry-aware focal cost (GFC), mask geometry-aware focal loss (MGFL) and mask geometry-aware focal cost (MGFC).}
    \vspace{-1mm}
    \label{tab:align}
    \end{table}

We further investigate the geometry-aware classification score $S_{geo}$. As depicted in Table~\ref{tab:aligntarget}, using $S_{giou}$, $S_{p2p}$ and $S_{dir}$ leads to individual improvement of -1.0, 2.8 and 4.7 mAP, respectively. Furthermore, the combination of $S_{p2p}$ and $S_{dir}$ boosts the performance, improving by 4.9 mAP.\@
\begin{table}[t]
    \centering
    \renewcommand\tabcolsep{7pt}
    \resizebox{0.47\textwidth}{!}{
    \begin{tabular}{@{}ccc|cccc@{}}
    \toprule
    $S_{giou}$ & $S_{p2p}$ & $S_{dir}$ & AP$_{ped}$ & AP$_{div}$ & AP$_{bou}$ & mAP \\
    \midrule
    & & & 51.7 & 58.7 & 59.4 & 56.6 \\
    \checkmark& & & 52.6 & 56.2 & 58.0 & 55.6 \\
    & \checkmark& & 54.8 & 60.2 & 63.1 & 59.4 \\
    & & \checkmark& 58.3 & 63.1 & 62.5 & 61.3 \\
    & \checkmark& \checkmark& 57.1 & 63.8 & 63.7 & 61.5 \\
    \bottomrule
    \end{tabular}
    }
    \caption{Ablation study of geometry-aware classification score.}
    \vspace{-4mm}
    \label{tab:aligntarget}
    \end{table}

\section{Conclusion}\label{sec:conclusion}

In this paper, we introduce InteractionMap, a novel approach to end-to-end online vectorized HD map utilizing temporal and spatial information interaction from local to global. By leveraging relation embedding at both point level and instance level, the network learns the geometric priors of map elements better. Subsequently, key-frame-based streaming strategy enhances the network by utilizing temporal feature embedding from near to distant. Finally, the geometric-aware focal loss and the geometric-aware focal cost build a strong correlation between classification score and instance-point location precision. Across various experimental settings, our proposed InteractionMap yields significant performance improvements over the baseline on various datasets.

{
    \small
    \bibliographystyle{ieeenat_fullname}
    \bibliography{main}
}

\clearpage
\setcounter{page}{1}
\maketitlesupplementary

In this supplementary file, we provide additional qualitative visual results of the proposed InteractionMap due to space limitation.

\setcounter{figure}{0}
\setcounter{section}{0}
\renewcommand\thesection{\Alph{section}}

\section{Qualitative Visualization}
\begin{figure*}[t]
    \begin{center}
    \centering
    \includegraphics[width=1.0\textwidth]{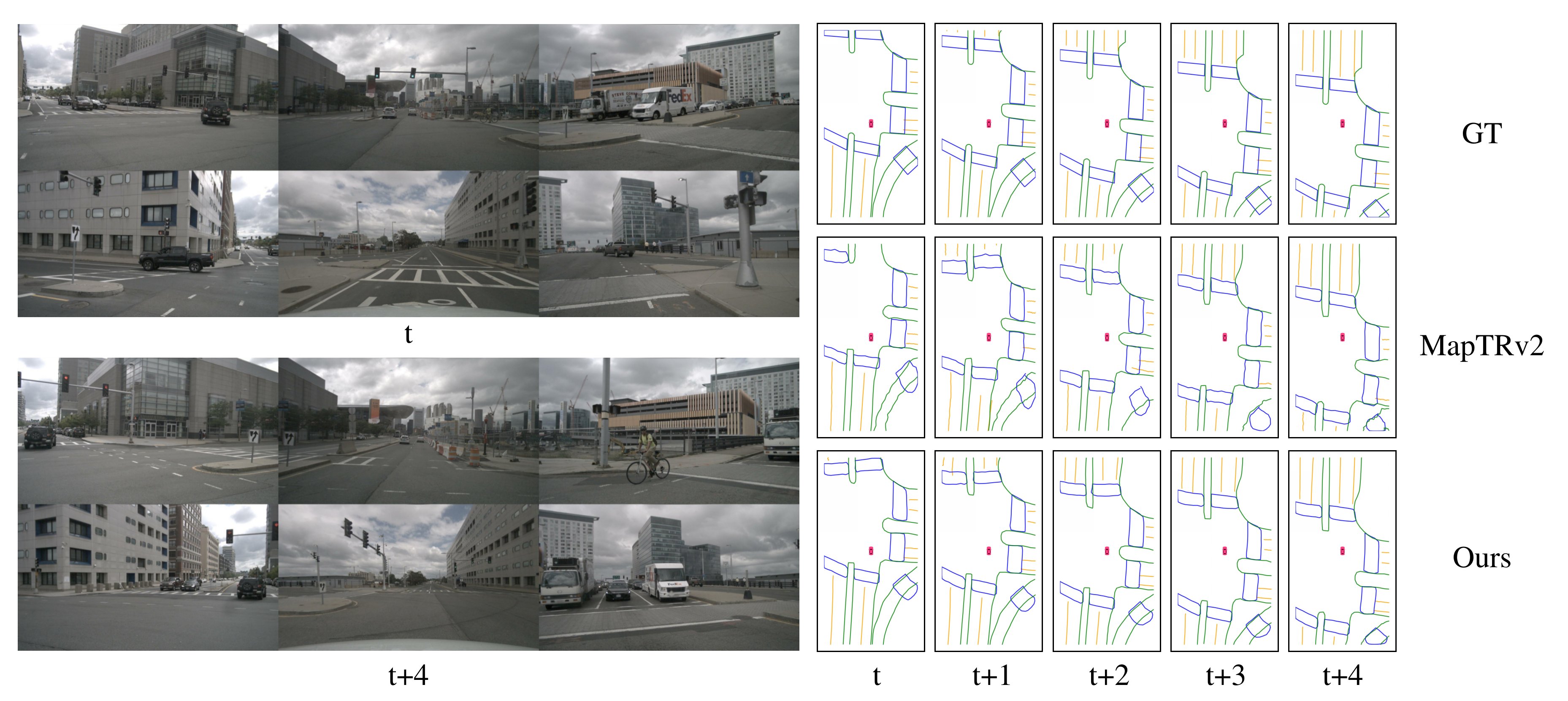}
    \caption{The visual results under the weather condition of cloudy.}
    \label{fig:cloudy1}
    \end{center}
    \end{figure*}

\begin{figure*}[t]
    \begin{center}
    \centering
    \includegraphics[width=1.0\textwidth]{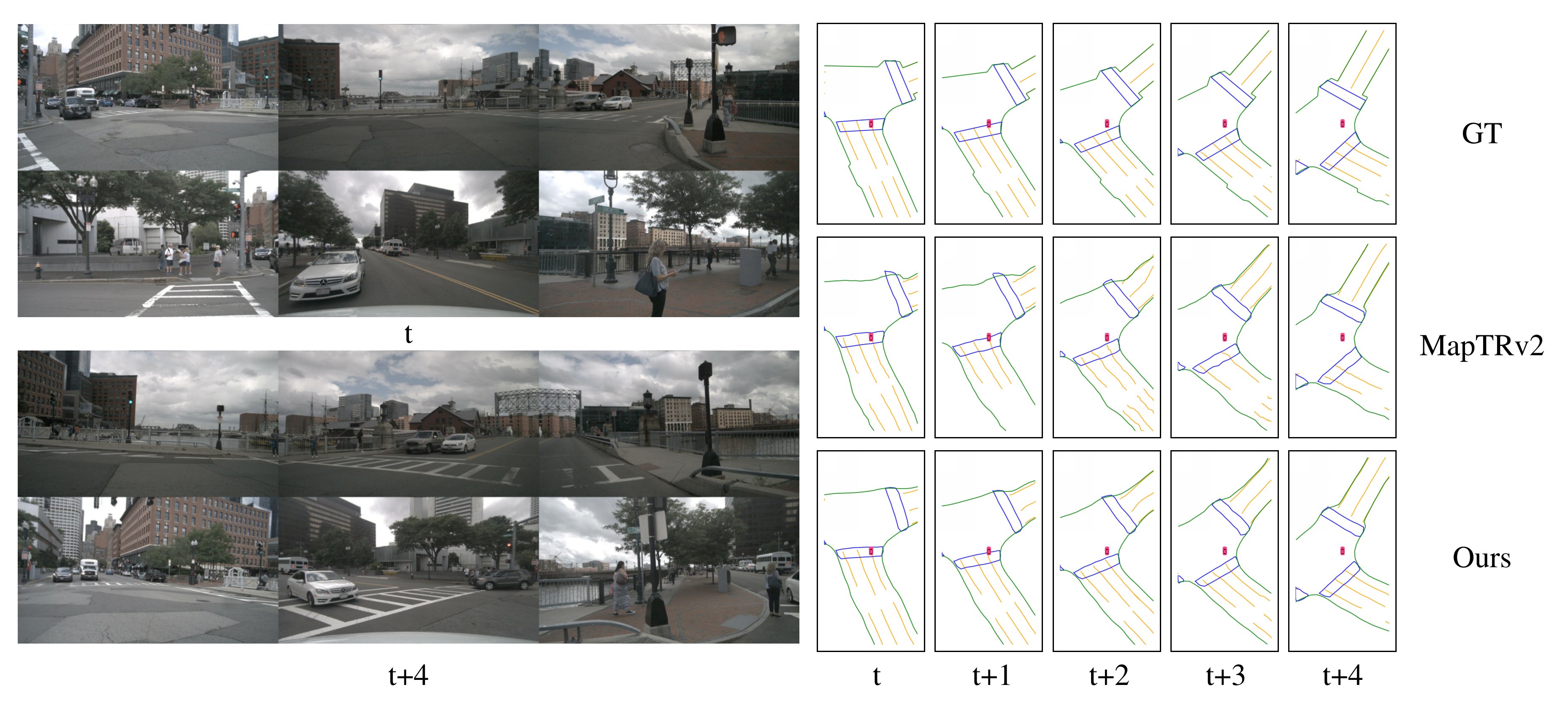}
    \caption{The visual results under the weather condition of cloudy.}
    \label{fig:cloudy2}
    \end{center}
    \end{figure*}

\begin{figure*}[t]
    \begin{center}
    \centering
    \includegraphics[width=1.0\textwidth]{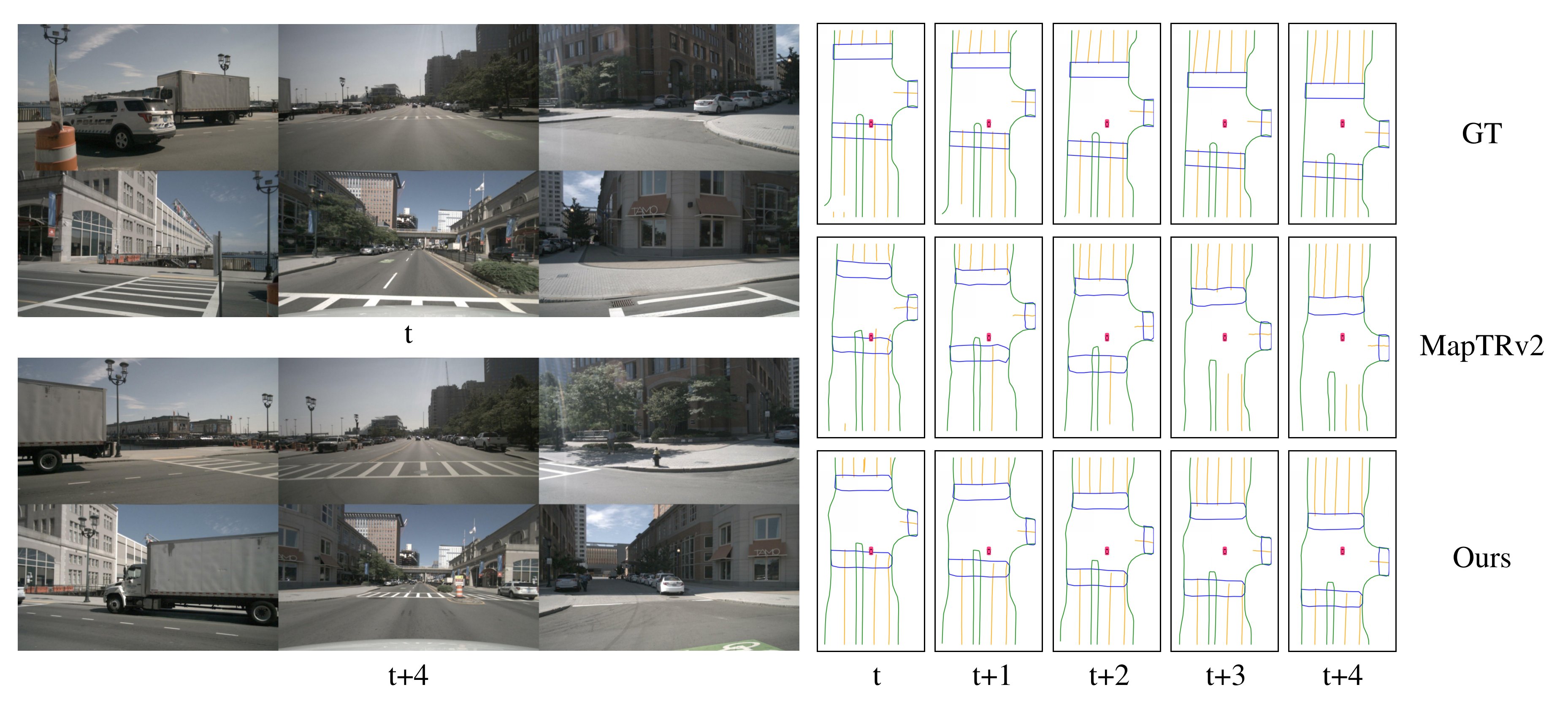}
    \caption{The visual results under the weather condition of sunny.}
    \label{fig:sunny1}
    \end{center}
    \end{figure*}

\begin{figure*}[t]
    \begin{center}
    \centering
    \includegraphics[width=1.0\textwidth]{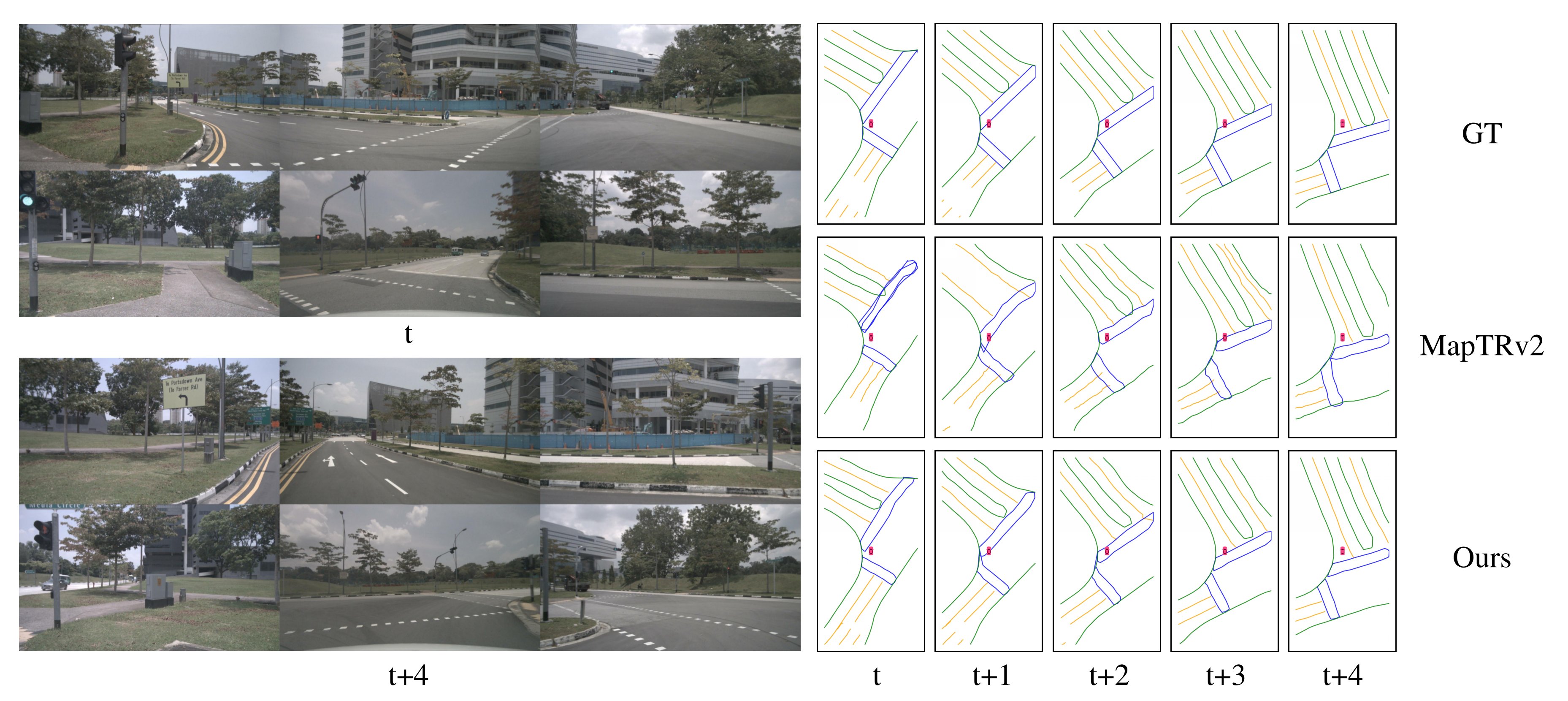}
    \caption{The visual results under the weather condition of sunny.}
    \label{fig:sunny2}
    \end{center}
    \end{figure*}

\begin{figure*}[t]
    \begin{center}
    \centering
    \includegraphics[width=1.0\textwidth]{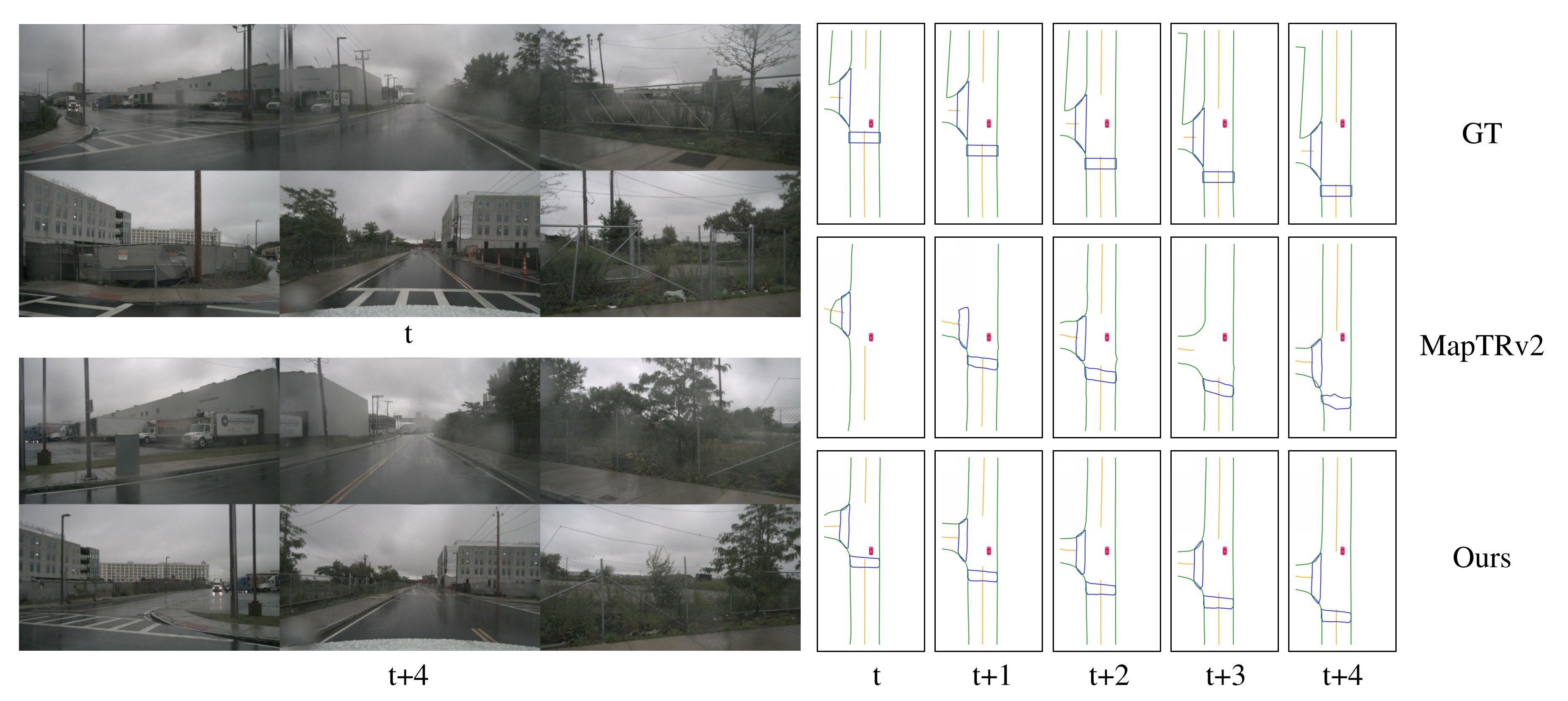}
    \caption{The visual results under the weather condition of rainy.}
    \label{fig:rainy1}
    \end{center}
    \end{figure*}

\begin{figure*}[t]
    \begin{center}
    \centering
    \includegraphics[width=1.0\textwidth]{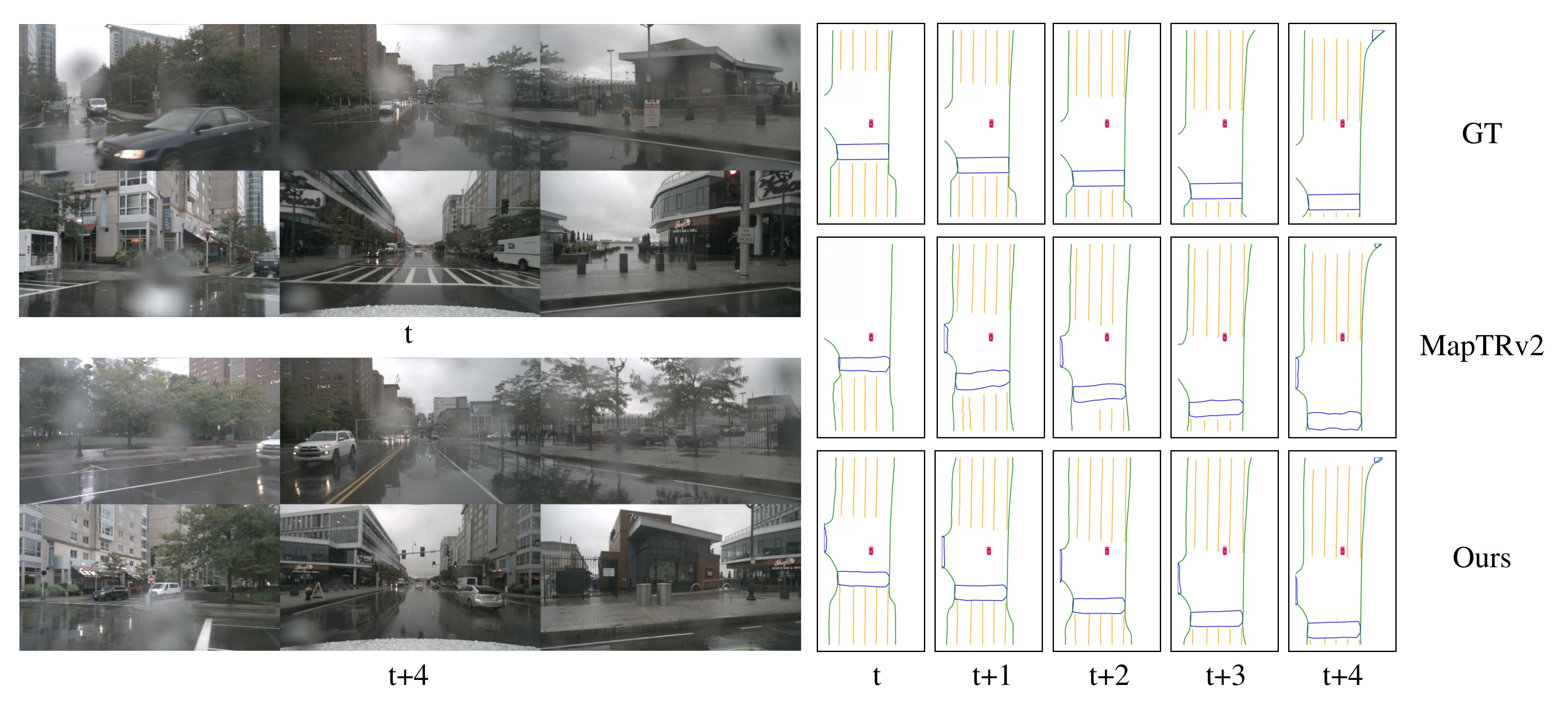}
    \caption{The visual results under the weather condition of rainy.}
    \label{fig:rainy2}
    \end{center}
    \end{figure*}

\begin{figure*}[t]
    \begin{center}
    \centering
    \includegraphics[width=1.0\textwidth]{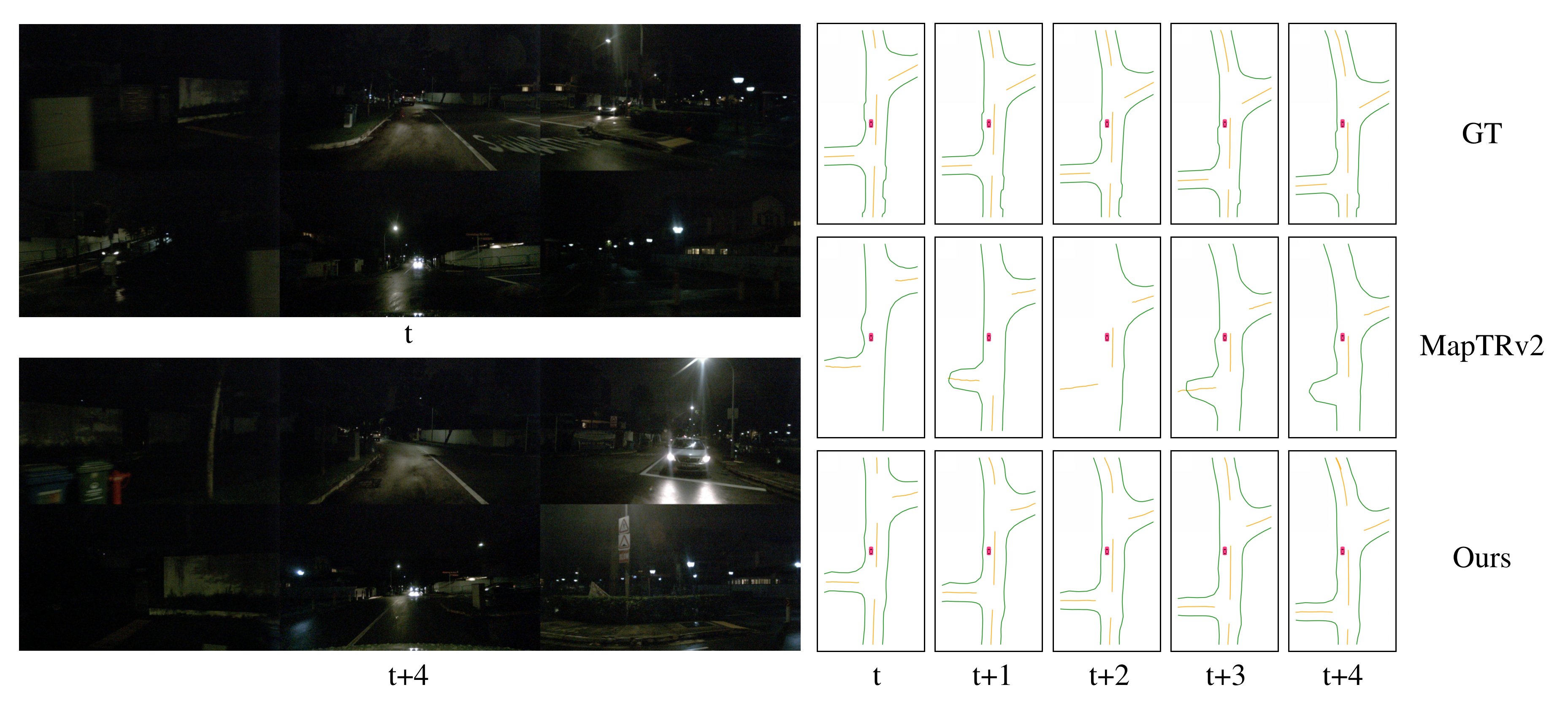}
    \caption{The visual results under the lighting condition of nighttime.}
    \label{fig:night1}
    \end{center}
    \end{figure*}

\begin{figure*}[t]
    \begin{center}
    \centering
    \includegraphics[width=1.0\textwidth]{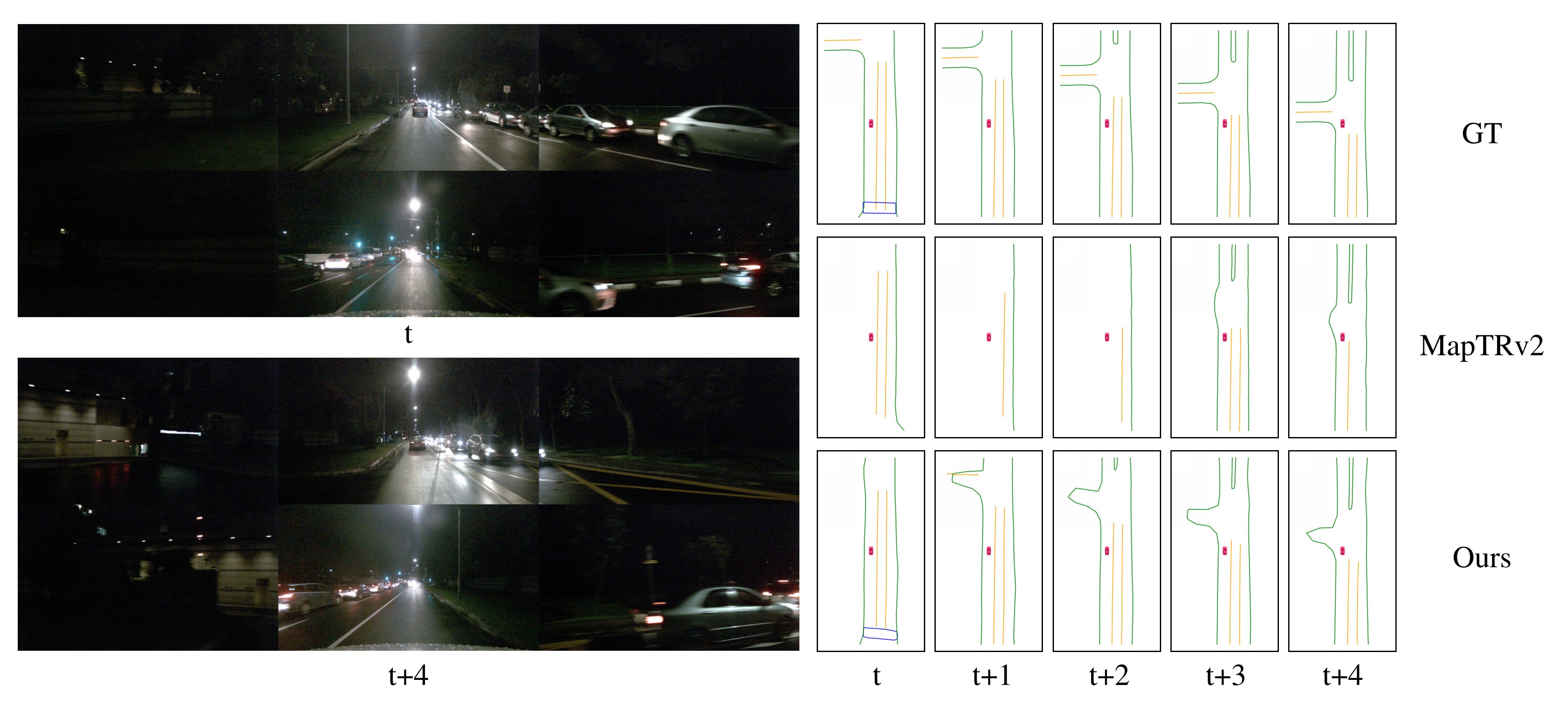}
    \caption{The visual results under the lighting condition of nighttime.}
    \label{fig:night2}
    \end{center}
    \end{figure*}

We visualize results of InteractionMap in sequential frames.
The visual results under the weather conditions of cloudy, sunny and rainy are shown in~\Cref{fig:cloudy1,fig:cloudy2},~\Cref{fig:sunny1,fig:sunny2},~\Cref{fig:rainy1,fig:rainy2}, respectively. And the visual results under the lighting condition of nighttime are shown in~\Cref{fig:night1,fig:night2}.

\end{document}